\documentclass{article} % options draft|referee|final
\usepackage{graphicx}
\usepackage[english]{babel}
\usepackage[utf8]{inputenc} % allow utf-8 input
\usepackage[OT1]{fontenc}    % use 8-bit T1 fonts
\usepackage{hyperref}       % hyperlinks
\usepackage{url}            % simple URL typesetting
\usepackage{booktabs}       % professional-quality tables
\usepackage{amsfonts}       % blackboard math symbols
\usepackage{nicefrac}       % compact symbols for 1/2, etc.
\usepackage{microtype}      % microtypography

\hypersetup{
    colorlinks=true,       % false: boxed links; true: colored links
    linkcolor=black,       % color of internal links
    citecolor=blue,       % color of links to bibliography
    filecolor=black,       % color of file links
    urlcolor=blue    % color of external links
}

\author{Ketil Malde and Hyeongji Kim \vspace{0.5cm} \\ 
Institute of Marine Research, PB-1507 Nordnes, Bergen, Norway \and Department of Informatics, University of Bergen, Norway}
\date{\today}
\title{Beyond image classification: zooplankton identification with deep vector space embeddings}
\hypersetup{
 pdfauthor={},
 pdftitle={Beyond classification: zooplankton identification with deep vector space embeddings},
 pdfkeywords={},
 pdfsubject={},
 pdfcreator={Emacs 25.2.2 (Org mode 9.1.6)}, 
 pdflang={English}}
\begin{document}

\maketitle
\begin{abstract}

Zooplankton images, like many other real world data types, have intrinsic
properties that make the design of effective classification systems
difficult.  For instance, the number of classes encountered in practical settings is potentially very
large, and classes can be ambiguous or overlap.  In addition, the choice of taxonomy
often differs between researchers and between institutions.  Although
high accuracy has been achieved in benchmarks using standard classifier
architectures, biases caused by an inflexible classification scheme can
have profound effects when the output is used in eco\-sys\-tem assessments
and monitoring.

Here, we propose using a deep convolutional network to construct a
vector embedding of zooplankton images. The system maps (embeds) each
image into a high-dimensional Euclidean space so that distances
between vectors reflect semantic relationships between images. 
We show that the embedding can be used to derive classifications with
comparable accuracy to a specific classifier, but that it
simultaneously reveals important structures in the data.  Furthermore,
we apply the embedding to new classes previously unseen by the system,
and evaluate its classification performance in such cases.

Traditional neural network classifiers perform well when the classes
are clearly defined \emph{a priori} and have sufficiently large labeled data sets available.  
For practical cases in ecology as well as in many other fields this is not the case, and
we argue that the vector embedding
method presented here is a more appropriate approach.

\end{abstract}

% \keywords{vector embedding, image classification, deep learning}

\section{Introduction}
\label{sec:orgffcb568}

In classification problems, the goal is to map each input to one of a
discrete set of classes.  A typical example is labeling images
according to objects pictured, e.g., distinguishing pictures of cats from
pictures of dogs.
The output of a classifier can be a single value, but is often a vector where
each element represents the classifier's confidence that the input
belongs to the corresponding class.

Recently, deep neural networks have been used with great success for
many classification tasks.  Often, these classifiers apply a softmax
function (a generalization of the logistic function to multiple
outputs) to generate the final output. This scales the output vector
so that the scores for the classes sum to one, resembling a set of
probabilities for the class assignment.

This approach is commonly used for image classification, where it has
been overwhelmingly successful for many benchmark data 
sets.  Yet, it relies on a set of assumptions that can
be naive in many practical situations.
Here, we will use the classification of zooplankton images 
to illustrate why a vector space embedding can be a more
appropriate approach.

\subsection{Zooplankton classification}
\label{sec:org35244e7}

Plankton constitute a fundamental component
of aquatic eco\-sys\-tems, and since they form the basis for many food
chains and also rapidly adapt to changes in the environment,
monitoring plankton diversity and abundances is a central input to marine
science and management \cite{ices2018wkmlearn}.

Imaging systems are
being deployed to scale up sampling efforts \cite{stemmann2012plankton,benfield2007rapid},
but the manual curation process remains expensive and time consuming \cite{ices2018wkmlearn}.
Recently, automated classifiers based on deep neural networks
have been developed and applied successfully to benchmark problem sets,
but deployment in a practical marine management
situation poses some challenges. 

For standard classifiers, the set of target classes is an integral
part of the structure of the classifier.  In other words, the set of
target classes must be finite and known in advance.  In contrast, plankton
communities often consists of surprisingly large numbers of species
(e.g., \cite{huisman1999biodiversity,schippers2001does}), with highly
varying abundance.  Even if all species were known, many would not
be represented in the training data, and the long-tailed abundance distribution
poses a challenge to standard methods \cite{van2017devil}.  A further complication is the
various forms of artifacts, including detritus, clusters of multiple specimens,
and pieces of fragile plankton that break apart during processing \cite{benfield2007rapid}.

In addition, different researchers may operate with
different taxonomies, or otherwise suffer from inconsistent annotation
\cite{10.1093/icesjms/fsz057}.  It is symptomatic that comparing the ZooScan data set used here
with another, similar data set \cite{DBLP:journals/corr/OrensteinBPS15}
with around 100 classes, we find that only three of the classes are
shared. Two of those represent artifacts  (\emph{bubble} and \emph{detritus}),
and only one plankton taxon (\emph{coscinodiscus}) was present as a
class in both data sets.
While it is possible to train classifiers separately for each
taxonomy, this diminishes the total value of the data and inhibits
comparisons and reproducibility.

Several automated systems for plankton classification have been
developed and applied to benchmark data sets
(e.g, \cite{luo2018automated,dai2016zooplanktonet,lee2016plankton}),
but report problems stemming from the severe class imbalance in the data.  In addition, image quality is often poor, and image sizes can
vary enormously.
In practice, automation is still mainly used to aid or supplement
a manual curation process \cite{uusitalo2016semi}.  For interactive processes, methods that
reveal more of the structure of the data are more useful than
categorical class assignments \cite{ices2018wkmlearn}.

\subsection{Vector embeddings as an alternative}
\label{sec:org42cb2b4}

Here we explore \emph{vector embedding} of the input space as an
alternative to the standard approach.  Each input is mapped to a
vector in a high-dimensional space with no \emph{a priori} relationship
between classes and dimensions.  Instead, the mapping (or embedding)
is constructed to reflect some concept of similarity between inputs.
In our case, class membership represents similarity, and the goal of
the embedding is to map inputs from the same class to vectors that are close to
each other, and inputs belonging to different classes to vectors that
are farther apart.

Compared to traditional classification, the embedding models
the structure of the input space with high resolution.  This is
important when the system deals with new classes of inputs.
Whether two inputs belong to the same or different classes can be
determined solely from the distance between their
corresponding vector space embeddings.  Similarly, new classes can be
constructed based on clusters or other structure in the embedding
vector space, without retraining or other modifications to the
system.

One application where neural networks that output embeddings have been applied with
particular success, is face recognition
\cite{taigman2014deepface,schroff2015facenet}.  Not unlike plankton
classification, the goal is to identify a large number of classes
(for face recognition, each individual person represent one class).  Thus we have a 
classification problem with an unknown, large, and possibly open-ended
number of classes, often with very sparse data and poor annotation.
As for face recognition, it is important to be able to identify
classes from few samples, so called \emph{low-shot}, \emph{one-shot} \cite{fei2006one}, and
\emph{zero-shot} \cite{larochelle2008zero,yu2010attribute} classification.

Inspired by this, we here apply a vector embedding approach to the task of classifying
zooplankton images, and compare the results to using a straightforward
classifier based on the Inception v3 \cite{szegedy2016rethinking} neural network architecture.  We
show how classes form clusters in the embedding space, discuss
confoundings, and explore how the vector embedding performs on
previously unseen classes.

\section{Methods}
\label{sec:orgba2ef40}

\subsection{Data set}
\label{sec:org25468d8}

Recently, a large set of ZooScan \cite{gorsky2010digital,grosjean2004enumeration} images of plankton was made available
to the public \cite{elineau2018}. 
The data set consists of monochromatic images organized into 93
categories, most of them representing zooplankton taxa.  In addition, several error
categories exist, with names like \emph{artefact}, \emph{detritus}, and
\emph{bubble}.  Abundances range from the 39 images labeled \emph{Ctenophora}, up to
the 511,700 labeled \emph{detritus}.  Three of the four most abundant
categories represent various types of artifact.

The images vary widely in size.  We converted the images to a standard
size of 299x299 pixels. Smaller images were padded up to
this size, while larger images were scaled down.
The resized images were then used to construct data sets for training,
validation, and testing.
For training, we used 65 non-artifact classes with abundances
above 500, in addition to \emph{bubble}.  From each class, 100 random images 
were selected to serve as a validation set, and then another
100 images for the test set.  The remaining images constituted the
training set.

A second test set consisted of 100 images sampled randomly from each
of the 38 classes not represented in the other sets.  For the classes
with less than 100 images, all images were used.

\subsection{Standard neural network classifier}
\label{sec:orgc4f7b56}

To provide a baseline for achievable classification accuracy, we used
the convolutional neural network Inception v3, initialized with
weights pre-trained on the ImageNet data set \cite{deng2009imagenet}.  The default 1000-class
output layer was replaced with a 65-class softmax output to match the
number of classes.

The network was trained using the SGD optimizer with a learning rate
of 0.0001 and momentum of 0.9, using a categorical cross-entropy cost
function.  During training, mean square error and accuracy were
reported.

All neural networks were implemented using Keras \cite{chollet2015keras} with a
Tensorflow \cite{abadi2016tensorflow} backend, and run on a computer with RTX2080 Ti
GPU accelerators (Nvidia Corporation, Santa Clara, California, USA).

\subsection{Siamese networks}
\label{sec:org8618c18}

The particular embedding technique we will investigate here is called \emph{siamese networks}
\cite{bromley1994signature,hoffer2015deep,wang2014learning}, in a variant
using what is called a triplet loss function.
The network is given three inputs, one from a randomly selected class (the
\emph{anchor}), one randomly sampled from the same class (the \emph{positive})
and a random sample from another class (the \emph{negative}).  The cost
function \(J\) is designed to reward a small distances from the anchor to
the positive and a large distance from the anchor to the negative.

\(J(a,p,n) = \max(0, ||a-p||^2 - ||a-n||^2 + \alpha)\)

The parameter \(\alpha\) serves as a margin to avoid the network
learning a trivial, zero-cost solution of embedding all inputs in the
same point.

For vector space embedding, we again used Inception v3, but replaced
the softmax with a global average pooling layer and a
128-dimensional vector output layer.  The output vector was further
constrained to unit length, so that the vector embedding results in a
point on a hypersphere with a radius of one.

Training was performed using the SGD optimizer and a batch size
of 20.  The learning rate was set to decay of 0.9 and an initial value
of 0.01.  The margin parameter \(\alpha\) was initially set to 1.0, but
raised to 1.3 after 20 iterations, and to 1.5 after 30 iterations.

\subsection{Classification from a vector space embedding}
\label{sec:org713ff43}

A vector space embedding does not directly present a classification,
but we can use any of a number of methods suitable for euclidean spaces.
An advantage of vector space embeddings is to allow the use of
unsupervised methods, and when no known data is available, classes
can be determined using standard approaches like \$k\$-means clustering.

Here, we will compare classifications in the embedding space using two
simple supervised methods.  First, using data with known
classes we calculate the centroids for each class and assign new
data to the class represented by the closest centroid.  Alternatively,
we use nearest neighbor classification (kNN, using the approximative
algorithm BallTree from Scikit-learn \cite{scikit-learn}) against data with known
classifications.  

\section{Results}
\label{sec:org1557b0d}

\subsection{Baseline classification}
\label{sec:org12cac80}

\begin{figure}[htbp]
\centering
\includegraphics[width=\linewidth]{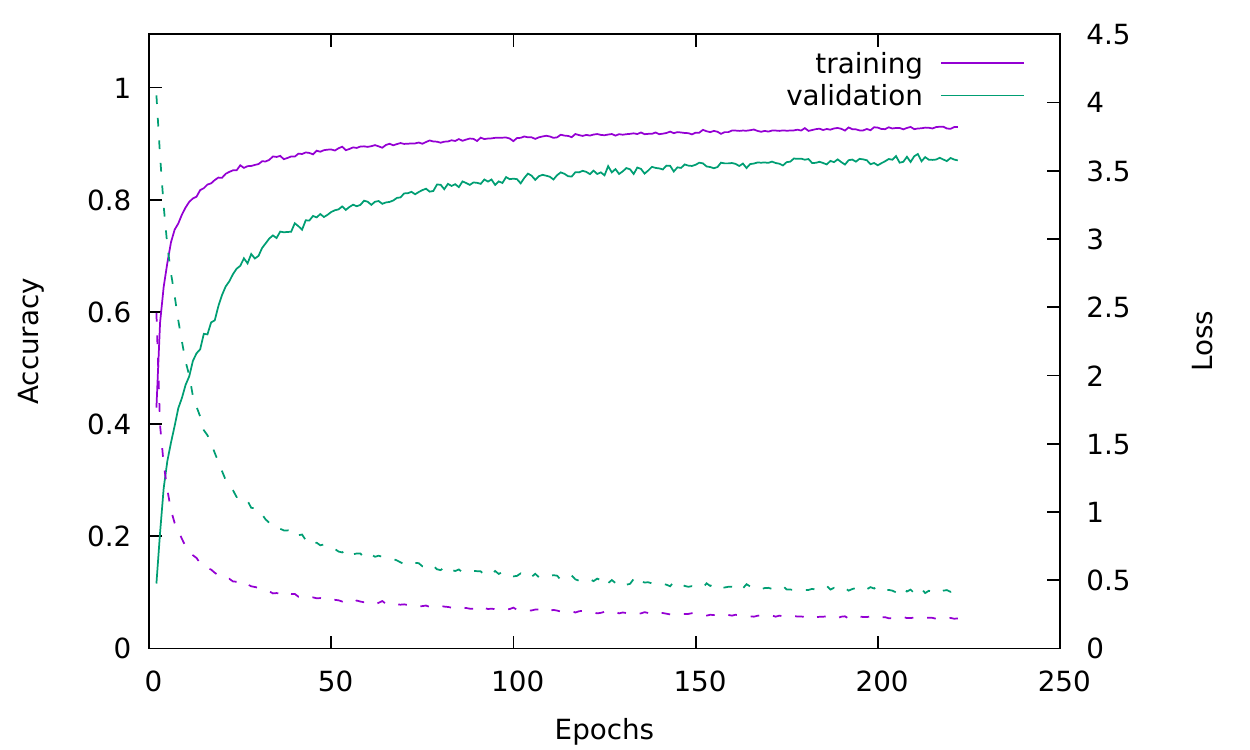}
\caption{\label{fig:orgd9cd1b9}
Training Inception v3 for plankton classification.  The loss (solid) and accuracy (dashed) for training (purple) and validation (green) data are shown as training progresses.}
\end{figure}

Inception v3 was trained for 220 epochs on the 65-class training data
set, the metrics are shown in Fig. \ref{fig:orgd9cd1b9}.  The classifier
reaches 80\% accuracy on validation data after 67 epochs, and appears
to converge to approximately 86\% accuracy after around 150 epochs.

We select the classifier trained for 200 epochs, and use it to
classify the test set.  Total accuracy was 87.7\%, a table with more
detailed results for the different classes can be found as
supplementary information.

\subsection{Training the vector embedding}
\label{sec:orgab49da1}

For validation, we calculated the centroid of the embeddings for each
category of plankton.  We define the cluster radius to be the average
distance from the centroid for each image in the validation set.
During training, we calculate the cluster radius (Suppl. Fig 1) and the change in
centroid (Suppl. Fig 2) for every class in the training set.  
As training progresses, cluster radii shrink, while the magnitude
of the changes to the embedding decreases.  In some cases,
large magnitude changes affect many or all clusters simultaneously,
indicating larger scale rearrangements in the embedding.

We can also check if we are able to correctly predict the correct
class by assigning each image to the closest centroid.  The results
are shown in Suppl Fig. 3.  Both analyses show rapid improvement for
10 iterations, slower gains the next 20, and only small improvements
after 30 iterations.  In the following, we use the network trained for
30 iterations to construct the vector embeddings.

\subsection{Clusters in the embedding space}
\label{sec:orgb7e0d54}

\begin{figure}[htbp]
\centering
\includegraphics[width=\linewidth]{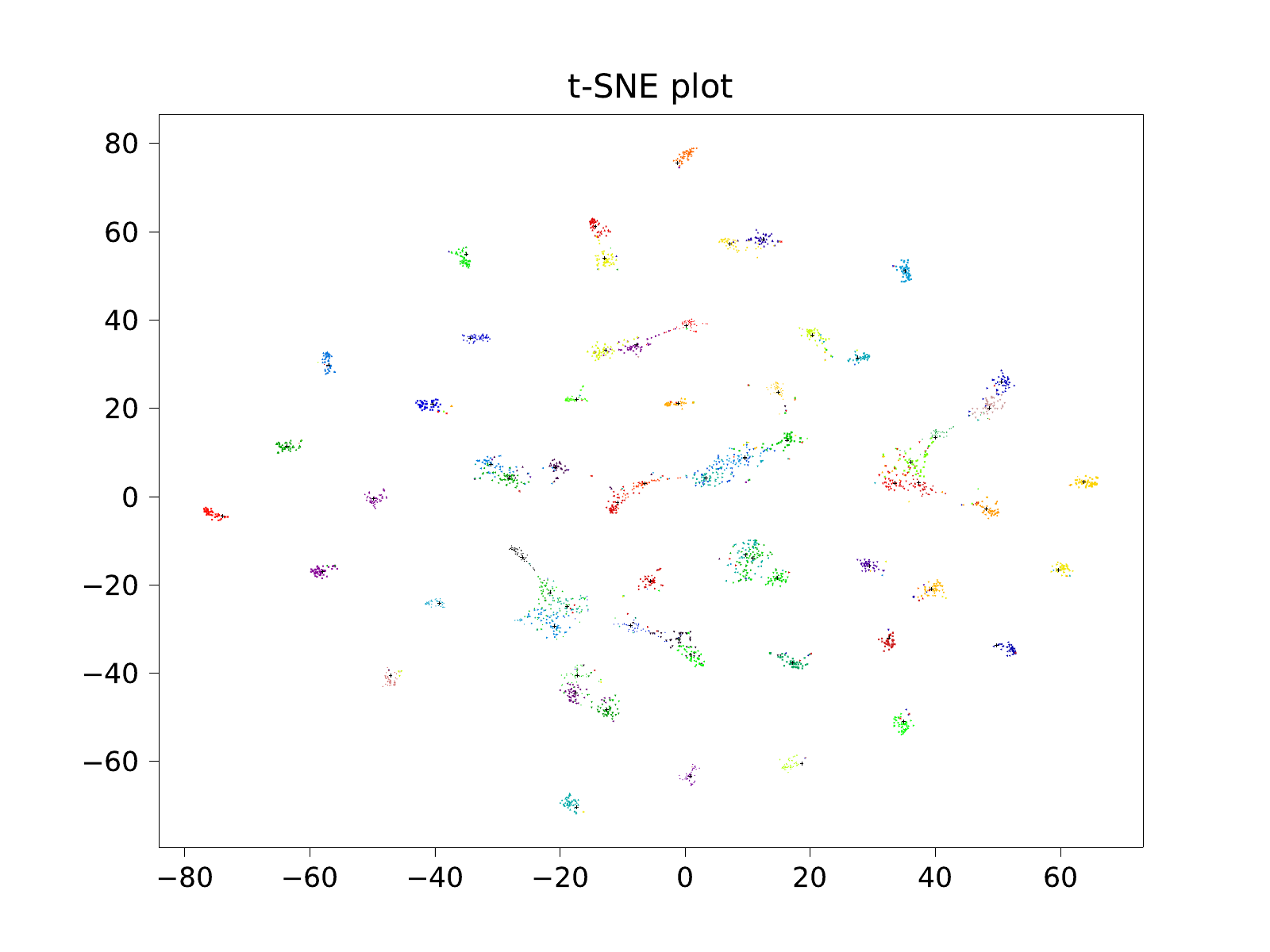}
\caption{\label{fig:org3d7e3ad}
The data projected into the embedding space and displayed using a t-SNE rendering.  Class centroids are marked by black crosses.}
\end{figure}

As training progresses, clusters start to emerge in the embedding
space.  A t-SNE \cite{maaten2008visualizing} rendering is shown in
Fig. \ref{fig:org3d7e3ad}, where the structure of the input data is evident.

\subsection{Classification in the embedding space}
\label{sec:org8efc766}

For classification using kNN, we investigate possible choices
for the parameter \(k\).  We split the validation data set in two (50 instances
for each class in each partition), and used one partition as a reference to classify
the other.  Experimenting with different values of  \(k\) indicates that
\(k=10\) might be a good value to use (see supplementary figure).

Fig. \ref{fig:org12e7ea2} shows the F1 scores using the default
classifier on the whole data set.  In addition, we show the centroid-based
classification in the embedded space and kNN classification using
various values of \(k\), splitting the test set into equal partitions
for reference and an evaluation.

We see that performance is comparable across most classes, but there
are some classes where the standard classifier gives different
performance from the embedding.  The standard classifier outperforms
the embedding for \emph{nauplii\_\_Cursacea} (class 65, F1 scores of 0.93 and
0.69) and \emph{nauplii\_\_Cirripedia} (class 12, F1 0.97 and 0.51).  A
substantial difference is also observed for \emph{egg\_\_Cavolinia\_inflexa}
(class 64, F1 0.93 and 0.33) and \emph{egg\_\_Actinopterygii} (class 17, F1 0.94
and 0.70).   In contrast, the embedding has better performance for 
\emph{Calanoida} (class 36, F1 0.50 and 0.96) and \emph{larvae\_\_Crustacea}
(class 36, F1 0.62 and 0.84).

\begin{figure}[htbp]
\centering
\includegraphics[width=\linewidth]{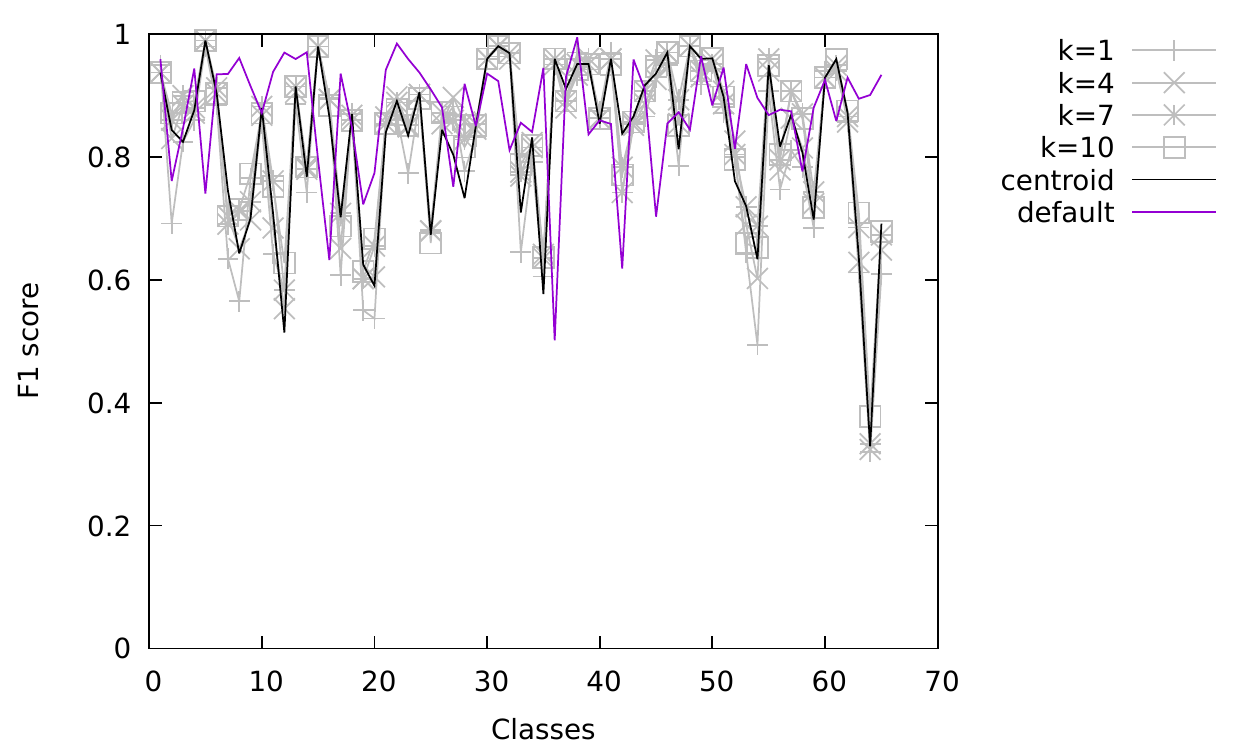}
\caption{\label{fig:org12e7ea2}
F1-scores across classes in the test set for centroid-based classification in black and kNN using various values for \(k\) in gray.  The standard classifier perfomance is shown in purple.}
\end{figure}

To elucidate the misclassifications, the ten most commonly occurring
confoundings with kNN (\(k=10\)) are shown in Table \ref{tab:orgbc7fe16}.

\begin{table}[htbp]
\caption{\label{tab:orgbc7fe16}
Commonly seen confoundings in the test data set.}
\centering
\small
\begin{tabular}{llr}
\hline
\textbf{True class} & \textbf{Predicted class} & \textbf{rate}\\
\hline
\emph{tail\_\_Appendicularia} & \emph{tail\_\_Chaetognatha} & 0.380\\
\emph{Oncaeidae} & \emph{Harpacticoida} & 0.260\\
\emph{Chaetognatha} & \emph{tail\_\_Chaetognatha} & 0.260\\
\emph{Euchaetidae} & \emph{Candaciidae} & 0.220\\
\emph{Eucalanidae} & \emph{Rhincalanidae} & 0.200\\
\emph{Harpacticoida} & \emph{Oncaeidae} & 0.180\\
\emph{nectophore\_\_Diphyidae} & \emph{gonophore\_\_Diphyidae} & 0.180\\
\emph{Rhincalanidae} & \emph{Eucalanidae} & 0.180\\
\emph{Centropagidae} & \emph{Euchaetidae} & 0.160\\
\emph{Limacidae} & \emph{Limacinidae} & 0.160\\
\hline
\end{tabular}
\end{table}

\begin{figure}[htbp]
\centering
\includegraphics[width=0.6\linewidth]{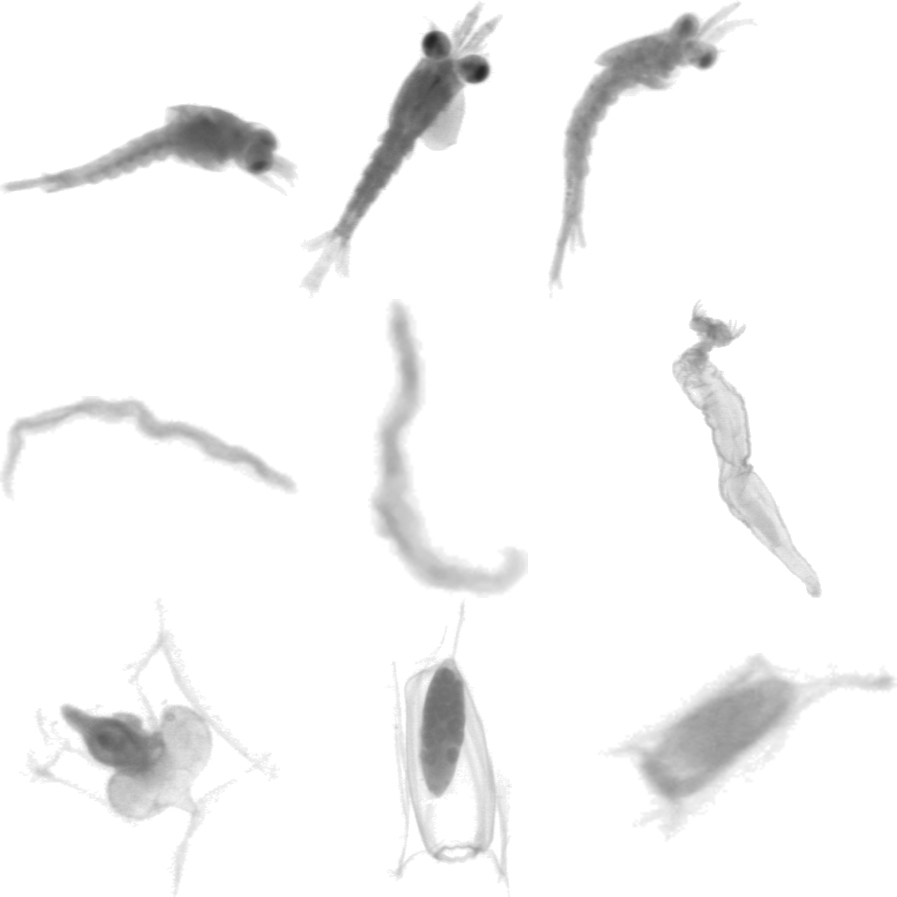}
\caption{\label{fig:orgfd30d56}
Example of plankton images that are difficult to resolve.  Upper row, from the left: \emph{Decapoda}, \emph{zooa\_\_Decapoda}, and \emph{larvae\_\_Crustacea}. Middle row: \emph{tail\_\_Appendicularia}, \emph{tail\_\_Chaetognatha}, and \emph{Chaetognatha}. Second row shows variants of \emph{Abylopsis}, from the left: \emph{eudoxie}, \emph{gonophore}, and \emph{nectophore}.}
\end{figure}

Not unexpectedly, confoundings occur between classes of
organism fragments or parts.  The most commonly occurring confounding consists of the two classes of
tails, and confounding \emph{Chaetognatha} with the class of its tails is
the third most common occurrence (see also Fig. \ref{fig:orgfd30d56}, middle row).
In addition, species are confounded with their different stages, e.g.,
we see confounding between different forms of the
\emph{Diphyidae} species (Fig. \ref{fig:orgfd30d56}, top row).

We also see pairs of similar species being confounded with each other
(e.g., \emph{Oncaeidae} with \emph{Harpacticoida}, and \emph{Eucalanidae} with
\emph{Rhincalanidae}).

\subsection{Previously unseen classes}
\label{sec:orgdaf349b}

\begin{figure}[htbp]
\centering
\includegraphics[width=\linewidth]{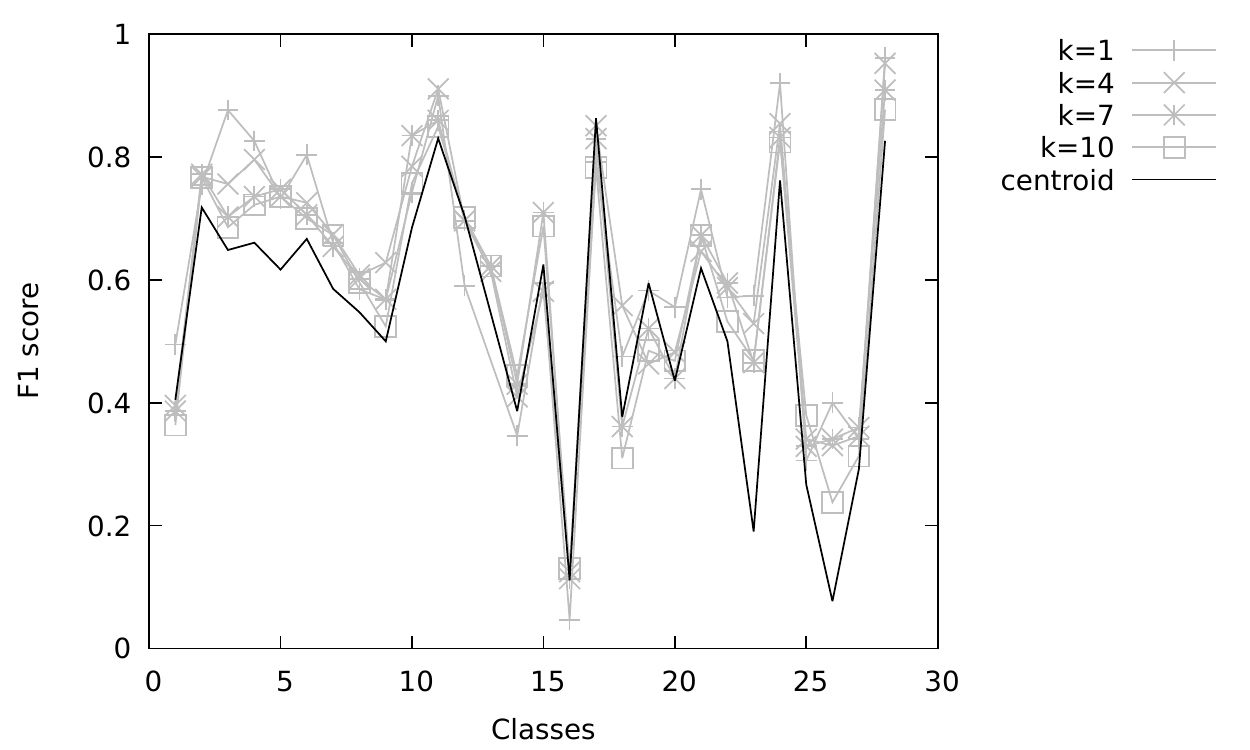}
\caption{\label{fig:org3494e35}
F1-scores across classes for centroid-based classification (black) and kNN using various values for \(k\) (gray).}
\end{figure}

For the previously unseen classes, we use the same approach of
dividing the test set in two and using one part for reference and the
other for evaluation.  The results are shown in Fig. \ref{fig:org12e7ea2}.
Here we see that performance is highly variable.  Using centroid
classification, the highest performing classes were Rhopalonema
(number 17, F1 0.86), badfocus\_artifact (number 28, F1 0.82), and
egg\_other (number 11, 0.83).  The lowest scoring classes were
Euchirella (number 26, F1 0.07), Aglaura (number 23, 0.19), and
multiple\_\_other (number 27, F1 0.29).  Several low performing classes
are caused by confusing the \emph{Abylopsis\_tetragona} variants (number 14, gonophore, F1 0.38, number 16
eudoxie, F1 0.11, and number 25, nectophore, F1 0.26).
kNN classifications outperforms
centroids slightly for several classes, but the overall picture
remains the same.

\begin{table}[htbp]
\caption{\label{tab:orgf707f04}
Commonly seen confoundings in previously unseen classes.}
\centering
\small
\begin{tabular}{llr}
\hline
\textbf{True class} & \textbf{Predicted class} & \textbf{rate}\\
\hline
\emph{eudoxie\_\_Abylopsis\_tetrag} & \emph{nectophore\_\_Abylopsis\_tet} & 0.320\\
\emph{Scyphozoa} & \emph{ephyra} & 0.280\\
\emph{Rhopalonema} & \emph{Aglaura} & 0.260\\
\emph{gonophore\_\_Abylopsis\_tetr} & \emph{nectophore\_\_Abylopsis\_tet} & 0.240\\
\emph{Calocalanus pavo} & \emph{Euchirella} & 0.240\\
\emph{nectophore\_\_Abylopsis\_tet} & \emph{eudoxie\_\_Abylopsis\_tetrag} & 0.220\\
\emph{badfocus\_\_artefact} & \emph{detritus} & 0.200\\
\emph{Calocalanus pavo} & \emph{part\_\_Copepoda} & 0.180\\
\emph{Echinoidea} & \emph{larvae\_\_Annelida} & 0.180\\
\emph{artefact} & \emph{badfocus\_\_artefact} & 0.180\\
\hline
\end{tabular}
\end{table}

Again we see that a large fraction of the confoundings occur between variants
of species, in particular \emph{Abylopsis tetragona} (Fig. \ref{fig:orgfd30d56},
bottom row).
In addition, there is several cases of confounding between artifact classes.

\begin{figure}[htbp]
\centering
\includegraphics[width=\linewidth]{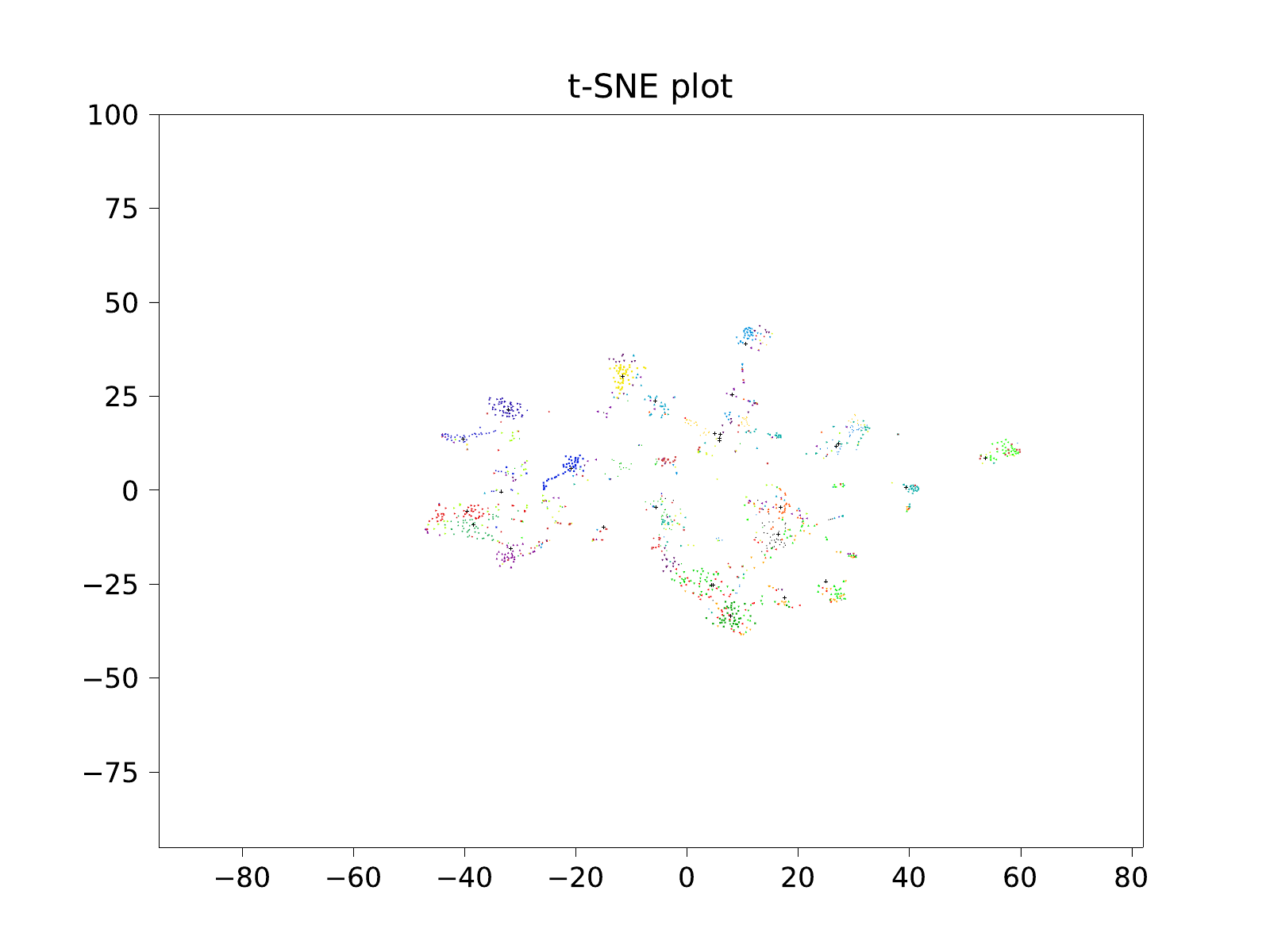}
\caption{\label{fig:orga02485d}
A t-SNE rendering of the data in the unseen classes after 30 iterations.}
\end{figure}

\section{Discussion}
\label{sec:org23dfd8c}

Using the average F1 score over the classes, the standard deep
learning classifier achieves a score of 0.87 on the data set. 
Using our vector space embedding and classifying using kNN (k=10), we
achieve a score of 0.84.  The standard classifier thus
outperforms the embedding, but not by a large margin.

Interestingly, the vector space embedding performs better on several
classes.  The standard classifier often mislabels many species as
\emph{Calanoida}, resulting in a low F1 score of 0.50, while the embedding
classifier achieves an F1 score of 0.96 for this class.  In contrast,
the standard classifier appears to be better at precisely separating
classes with very similar morphology, for instance classes of eggs or
nauplii.  As similar classes are embedded close to each other, they
are more difficult to differentiate.  Although the proximity is
semantically meaningful, this reduces accuracy somewhat. For
maximizing absolute classification performance, an ensemble using both
methods is likely to be optimal.

In contrast to classification, the embedding is able to better capture
the underlying structure of the data.  This has many potential uses,
for instance to identify misclassified data, or to allow switching to
a different taxonomy.  In this way, the embedding can be used actively
to evaluate and even refine the choice of classes used.

As a more challenging test case, we applied the embedding approach to
data in classes not present in the training data.  Here we achieve a
more modest performance, with an average F1 score of 0.61.  Some of
the classes gave particularly poor results, while other classes were
accurately identified.  Even for classes where performance is too low
to be used directly, the information provided by the embedding can guide and
accelerate manual or semi-interactive processing.  We believe training
with more diverse data is likely to improve generality of the embedding.

The use of very simple schemes used to 
compare classification performance in the embedding space (\emph{i.e.}, centroid
clustering and kNN) is a deliberate choice.
More complex schemes may be able to give better classification
performance, but our goal here is to emphasize the ability of the
embedding to capture the structure of the input.  Using a complex
non-linear classifier on the embedding vectors would defeat this
purpose, since it would be more difficult to separate complexity captured
by the embedding from complexity captured by the final classification
stage.

\section{Conclusions}
\label{sec:org3c4b7bc}

Classification of zooplankton is an important task, but the inherent
complexity and other limitations of the data requires more flexibility
than that provided by standard classifiers.  Earlier attempts have
successfully been able to classify benchmark data sets \cite{py2016plankton,lee2016plankton}, but achieve
high accuracy at the expense of removing low abundance or otherwise
difficult classes \cite{luo2018automated}.  

Here we have shown that using a deep learning vector space embedding,
we can model important structure in the data, while retaining the
flexibility to perform classification with accuracy comparable to
state of the art classifiers. 

\section{Author's contributions}
\label{sec:orgfefa729}

KM conceived of the ideas and methodology and led the writing of the
manuscript, HK implemented benchmarks and visualizations of the
results.  Both authors contributed critically to the draft and
approved of its publication.

\section{Availability}
\label{sec:orgd3fe9c9}

The data set and software used here is publicly available as described above.  Source code
for network construction, training, and analysis can be found as GitHub repositories at

\url{https://github.com/ketil-malde/plankton-siamese} and
\url{https://github.com/ketil-malde/plankton-learn}

An interactive rendering of the data sets and classifications using \url{https://projector.tensorflow.org}/ can be found here:

\url{https://projector.tensorflow.org/?config=https://home.malde.org/vector\_embeddings/}

\small
\bibliographystyle{apalike}
\bibliography{references}
\end{document}

\begin{figure}[htbp]
\centering
\includegraphics[width=0.6\linewidth]{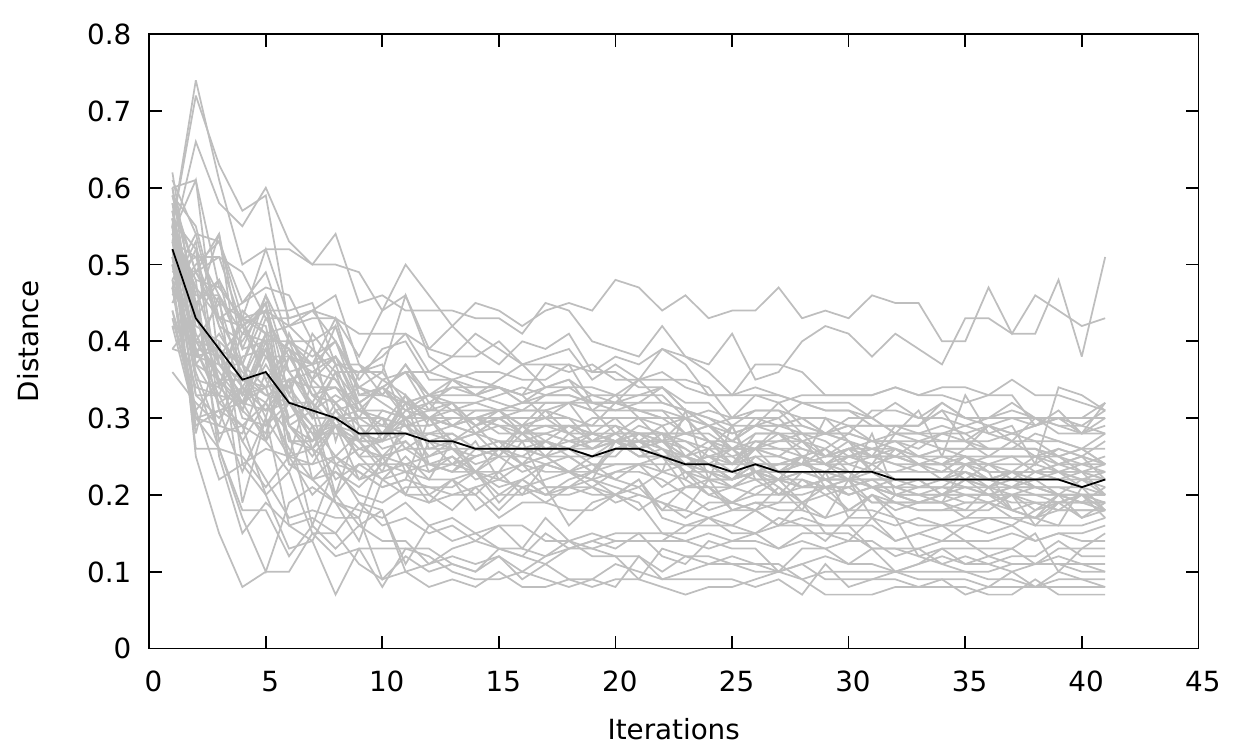}
\caption{\label{fig:orgc25ca3b}
Cluster radius for the 65 individual categories (gray) and the average (black) as training progresses.}
\end{figure}

\begin{figure}[htbp]
\centering
\includegraphics[width=0.6\linewidth]{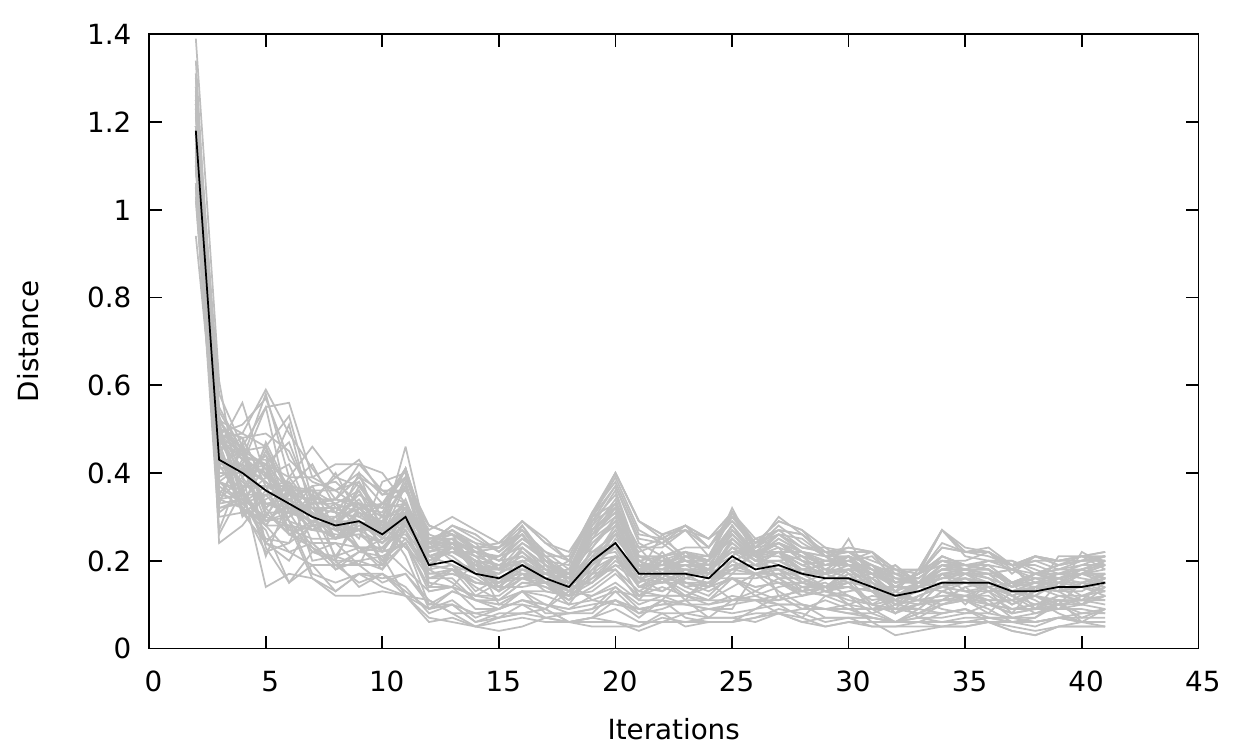}
\caption{\label{fig:org02c6c6b}
Output change measured as the distance cluster centroids move between iterations.  Individual centroids are shown in gray and the average in black.}
\end{figure}

\begin{figure}[htbp]
\centering
\includegraphics[width=0.6\linewidth]{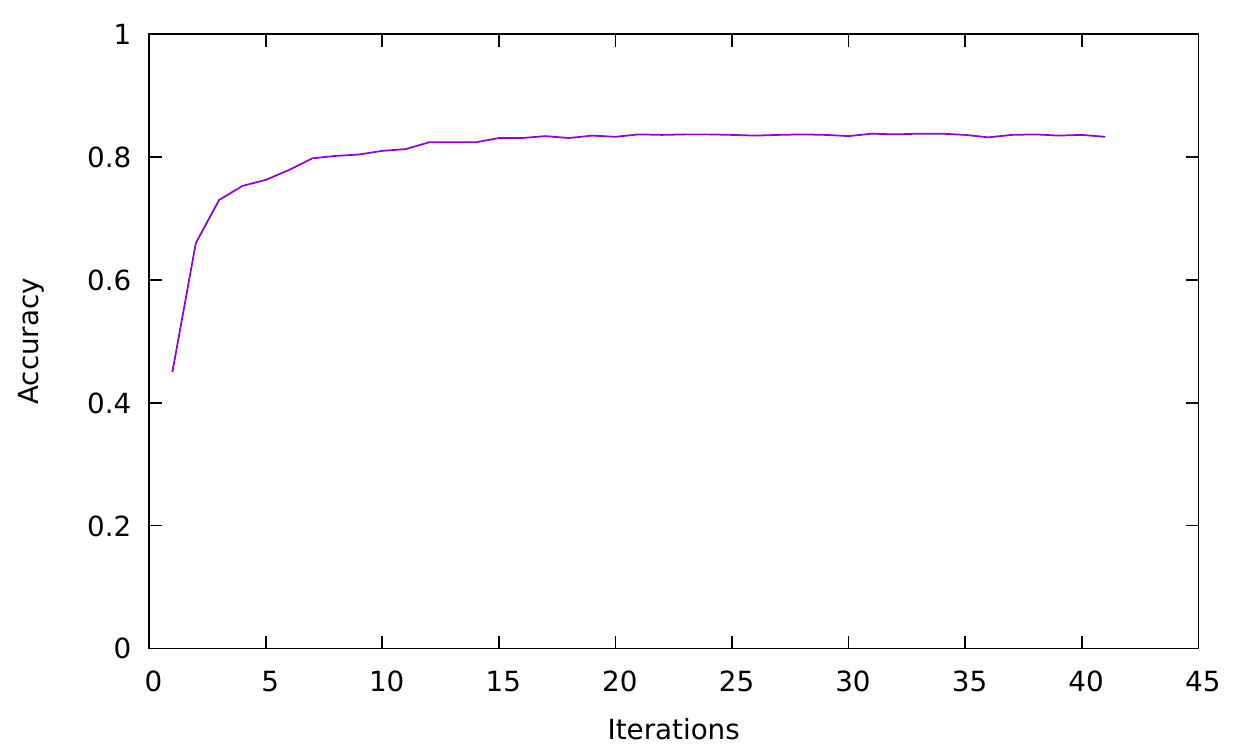}
\caption{\label{fig:orgad41be3}
Prediction accuracy from assigning each image to the nearest centroid.  Only the validation set is shown. Accuracy plateaus at 0.838 after 30 iterations, and decreases slightly after 35.}
\end{figure}

\begin{figure}[htbp]
\centering
\includegraphics[width=0.6\linewidth]{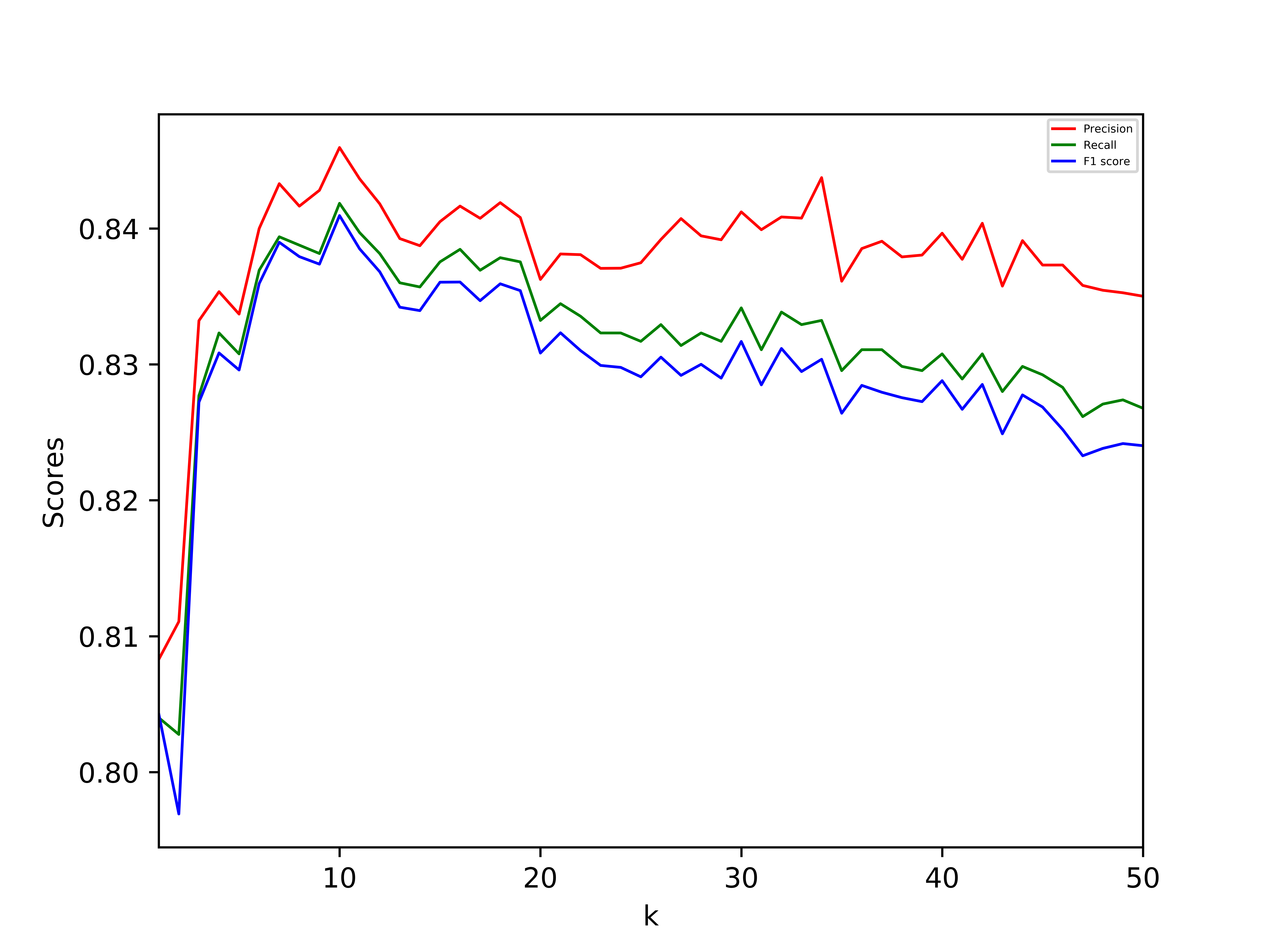}
\caption{\label{fig:org74b1661}
Classification performance using the kNN algorithm for different values of \(k\).}
\end{figure}

\newpage

\begin{table}[htbp]
\caption{\label{tab:org8ed8974}
Accuracy per class on the test set using Inception v3 for classification.  Confounders are the most frequent incorrect labels reported when the error occurs in more than 5\% of the cases.}
\centering
\scriptsize
\begin{tabular}{lrlllrl}
\textbf{Class} & \textbf{Recall} & \textbf{Confounders} &  & \textbf{Class} & \textbf{Recall} & \textbf{Confounders}\\
\hline
Acantharea & 82 & Phaeodaria (12) &  & Harpacticoida & 86 & Oncaeidae (8)\\
Acartiidae & 93 & Calanoida (7) &  & Hyperiidea & 76 & Calanoida (7)\\
Actinopterygii & 95 &  &  & larvae\_\_Crustacea & 47 & calyptopsis (22)\\
Annelida & 84 &  &  & Limacidae & 58 & Limacinidae (36)\\
Bivalvia\_\_Mollusca & 93 &  &  & Limacinidae & 87 & \\
Brachyura & 98 &  &  & Luciferidae & 72 & Decapoda (15)\\
bubble & 97 &  &  & megalopa & 94 & \\
Calanidae & 74 & Calanoida (24) &  & multiple\_\_Copepoda & 79 & Calanoida (7)\\
Calanoida & 97 &  &  & nauplii\_\_Cirripedia & 97 & \\
calyptopsis & 91 &  &  & nauplii\_\_Crustacea & 99 & \\
Candaciidae & 74 & Calanoida (16) &  & nectophore\_\_Diphyidae & 88 & gonophore\_\_Diphyidae (9)\\
Cavoliniidae & 91 &  &  & nectophore\_\_Physonectae & 94 & \\
Centropagidae & 60 & Calanoida (39) &  & Neoceratium & 90 & seaweed (6)\\
Chaetognatha & 96 &  &  & Noctiluca & 97 & \\
tail\_\_Chaetognatha & 39 &  &  & Obelia & 95 & \\
Copilia & 92 &  &  & Oikopleuridae & 99 & \\
Corycaeidae & 94 &  &  & Oithonidae & 95 & Calanoida (5)\\
Coscinodiscus & 99 &  &  & Oncaeidae & 89 & Corycaeidae (5)\\
Creseidae & 95 &  &  & Ophiuroidea & 93 & \\
cyphonaute & 100 &  &  & Ostracoda & 94 & \\
cypris & 81 & Ostracoda (13) &  & Penilia & 99 & \\
Decapoda & 92 &  &  & Phaeodaria & 95 & Foraminifera (5)\\
zoea\_\_Decapoda & 6 &  &  & Podon & 76 & Evadne (17)\\
Doliolida & 96 &  &  & Pontellidae & 94 & \\
egg\_\_Actinopterygii & 95 &  &  & Rhincalanidae & 77 & Eucalanidae (21)\\
egg\_\_Cavolinia\_inflexa & 82 & egg\_\_Actinopterygii (5) &  & Salpida & 92 & \\
Eucalanidae & 78 & Calanoida (9) &  & Sapphirinidae & 85 & \\
Euchaetidae & 65 & Calanoida (32) &  & scale & 82 & Noctiluca (5)\\
eudoxie\_\_Diphyidae & 80 & nectophore\_\_Diphyidae (13) &  & seaweed & 86 & \\
Evadne & 95 &  &  & tail\_\_Appendicularia & 85 & Chaetognatha (5)\\
Foraminifera & 87 &  &  & tail\_\_Chaetognatha & 49 & Chaetognatha (39)\\
Fritillariidae & 87 & Oikopleuridae (10) &  & Temoridae & 97 & \\
gonophore\_\_Diphyidae & 84 & nectophore\_\_Diphyidae (12) &  & zoea\_\_Decapoda & 92 & Decapoda (6)\\
Haloptilus & 91 & Calanoida (5) &  &  &  & \\
\end{tabular}
\end{table}